\documentclass{article}

\usepackage[final]{neurips_2023}

\usepackage[utf8]{inputenc} 
\usepackage[T1]{fontenc}    
\usepackage{hyperref}       
\usepackage{url}            
\usepackage{booktabs}       
\usepackage{amsfonts}       
\usepackage{nicefrac}       
\usepackage{microtype}      
\usepackage{xcolor}         
\usepackage{ragged2e}
\usepackage{graphicx}
\usepackage{amsmath}
\usepackage{amsfonts}
\usepackage{amssymb}

\title{Systematic Evaluation of Robustness in Fine-tuned Models }

\author{%
  Angela John  \\
  Saarland Informatics Campus, Saarland University\\
  \texttt{ajohn@cs.uni-saarland.de} \\
  \And
  Hamayoon Behmanush \\
  Saarland Informatics Campus, Saarland University \\
  \texttt{behmanush@cs.uni-saarland.de} \\
}

\begin{document}

\maketitle

\begin{abstract}
 Our study systematically evaluates the robustness of fine-tuned pre-trained models across different datasets and attack types, aiming to provide insights for researchers in selecting pre-trained models for image classification tasks while prioritizing robustness against potential attacks. Our work explores the influence of fine-tuning pre-trained models with different datasets and attack types on their robustness and overview of the appropriate pre-trained model for image classification based on factors such as computation resources, memory usage, and accuracy on clean data. We fine-tune the models with weights. Evaluate the models on data before any attack and after different attack cases. The study utilizes publicly available image datasets such as MNIST, CIFAR10, CIFAR100, and SVHN to assess the robustness of pre-trained models. The results indicate that the robustness of fine-tuned models varies across different data distributions and sample sizes, emphasizing the need for tailored selection based on specific task requirements. The findings also highlight the varying performance of ResNet50, DenseNet, and MobileNet in defending against attacks on different datasets. Overall, this research provides valuable insights into the robustness of fine-tuned pre-trained models and their suitability for real-world applications in image classification.
\end{abstract}

\section{Introduction}
The emergence of pre-trained models has revolutionized the field of machine/deep learning by enabling their application across diverse contexts and reducing computation costs. Leveraging these models can reduce training and computational resources\cite{sagemaker_blog_2022} by fine-tuning them for specific tasks. On the other hand, the robustness of any machine/deep learning model is crucial, and the performance models on noisy data may depend on how robust the model is \cite{taly_expect_unexpected}. Ensuring robustness often involves augmenting training data, incorporating regularization techniques, and deploying ensemble methods \cite{huang2021robustness}, not forgetting to account for the adversarial uncertainties expected. Moreover, continuous monitoring and adaptation of models to evolving data distributions are essential for maintaining robust performance in dynamic environments. Therefore, while pre-trained models offer significant advantages, ensuring their robustness when fine-tuning them remains an open challenge. \\
Potential research in the past, such as \cite{ban2022pre}, has explored the improvement of attack success rates with pre-trained perturbations. Furthermore, Dewan et al. \cite{dewan2023image} in their work performed a comparative analysis of pre-trained models to evaluate the performance of these models with diverse tasks. Despite these advancements, a comprehensive evaluation of the robustness of fine-tuned models across a range of attacks and different distribution datasets is an area that is yet to be fully covered and therefore worth to be explored. \\
This evaluation is crucial for understanding the model's defenses' generalizability and effectiveness in real-world scenarios. Additionally, investigating the transferability of adversarial examples across different fine-tuned models could provide insights into their underlying vulnerabilities and strengths. Such insights can inform the development of more robust and reliable deep learning systems for various applications, from image classification and beyond. With this in mind, we try to answer the following questions in this project:

\begin{itemize}
    \item How does fine-tuning pre-trained models with different datasets considering different adversarial perturbations, influence their robustness?
     \item Does the choice of attacks and epsilon values in attacks influence the adversarial accuracy?
    \item Is it possible to decide on a robust pre-trained model if the data distribution and size are not known?
     
\end{itemize}

\section{Related-Work}
\subsection{Rise of Pre-Trained Models}
Pre–trained models reduce the computational cost of training models from scratch and can be adapted to work with various datasets after fine–tuning. In recent years, adversarial perturbation and self-supervised pre-training domains, have received increasing attention despite the security concerns with the robustness of pre-trained models. Ban and Dong
\cite{ban2022pre} explain the robustness of pre-trained models by introducing pre-trained perturbations to improve the attack success rate but generalization on perturbed models is yet to be covered. Shi et al. \cite{shi2022robust} On the other hand, explain that self-supervised learning (SSL) models are more robust to distribution shifts compared to auto-encoder-based models (AE) and supervised learning (SL) models. The evaluation of real-world and synthetic distribution datasets like MNIST-CIFAR and CdSprites has been studied. Shao et al. \cite{zhao2022investigating} further explores the use of ensemble methods to improve the robustness of text classifiers, highlighting the importance of choosing an appropriate bias model. Hendricks et al. \cite{hendrycks2019using} emphasized the role of pre-training in improving model robustness and uncertainty estimates, particularly in the context of adversarial examples, label corruption, class imbalance, and out-of-distribution detection. Suya\cite{suya2023distributions} also demonstrates that linear learners can be robust to indiscriminate poisoning if the class-wise data distributions are well-separated with low variance and the size of the constraint set containing all permissible poisoning.
\subsection{Robustness of the Pre-training to Fine-Tuned Models}

Fine-tuning, a widely used approach for transfer learning from pre-trained models to new ones, involves adjusting parameters to tailor the model to specific tasks \cite{li2021improved}. These pre-trained models are initially trained on vast, diverse datasets to grasp general representations of the world. However, fine-tuning allows them to adapt to new, narrower tasks with smaller datasets. Yet, a crucial concern arises in ensuring that the fine-tuned model remains robust in the face of noise and potential perturbations, safeguarding its reliability and effectiveness in real-world applications. \\
Recent studies such  as \cite{zhu2023improving} have evaluated the performance of adversary pre-trained models by introducing  RiFT (Robustness Critical Fine-Tuning) while fine-tuning. Their findings underscore a notable correlation between generalization, adversarial robustness, and out-of-distribution robustness in fine-tuned models. Neerudu at el \cite{neerudu2023on} illustrates the pivotal role of pre-training data quality in determining a model’s capacity to be generalized in new novel tasks which highlights the importance of the quality of pre-trained data in the robustness of fine-tuned models. When fine-tuning the pre-trained models, size, quality, and distribution of data for fine-tuned models have a significant role \cite{xu2023neighborhood}. \\
These findings suggest that maintaining the robustness of fine-tuned models relies on various factors. However, there is still an ongoing challenge in systematically evaluating the robustness of fine-tuned models against different attacks and datasets to assist researchers in selecting the appropriate pre-trained model for fine-tuning.

\section{Methodology}
Recent studies have explored the use of different pre-trained models for image classification. For instance \cite{dewan2023image}, compared the performance of VGG16, Densenet, Mobilenet, Inception V3, Resnet50, and Xception models, although this has not been tested and compared in cases of adversarial perturbations to determine the most robust model. In this study, we intend to cover the models studied in \cite{dewan2023image}  under adversarial perturbations to evaluate and see if the models are still robust under adversarial uncertainties.

\begin{figure}[!h]
    \centering
    \includegraphics[width=0.9\textwidth]{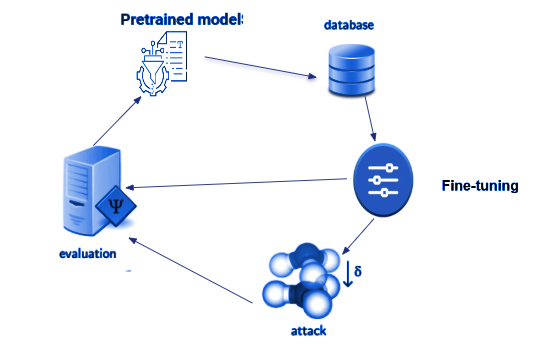} 
    \caption{Pipeline of our Approach}
    \label{pipeline}
\end{figure}

We fine-tune our models with weights and run 60 epochs for clean data, then evaluate the models with different distributions of clean datasets that we fine-tune with. We apply the attack to the models with epsilon values of between 0.001 and 0.5 and evaluate the accuracy of the datasets. These will give us an understanding of the limitations of using pre-trained models, whether we can still generalize under different distributions, and how well they can perform in the case of adversarial uncertainties. We adopt the pipeline as illustrated in \ref{pipeline}.We take our pre-trained model and fine-tune it with the datasets,on one direction we evaluate the fine-tuned model with the dataset. We also take the fine-tuned models expose it to the 3 different attacks and finally evaluate the model.
  \begin{center}
    
  \end{center}
\section{Experiments}
\subsection{Pretrained-Models}
In our analysis we adopt pre-trained models Resnet50, Densenet121, and Mobilenet-V2  pre-trained on the ImageNet datasets.\\
\textbf{ResNet50} boasts a 50-layer architecture with innovative "skip connections." These connections allow training deeper networks without  worrying about vanishing gradients \cite{he2016deep}. They tackle the vanishing gradient problem, a major hurdle in deep learning. ResNet50 achieves this by directly feeding high-resolution outputs from earlier layers to deeper layers, which initially handle lower-resolution information. These skip connections ensure the output is at least as good as the input from previous layers, minimizing the vanishing gradient effect during backpropagation. This powerful technique allows training deeper models while keeping loss levels low.\\
\textbf{Densenet-121} builds upon the idea of skip connections but takes it a step further \cite{huang2017densely}. Instead of just jumping to later layers, it passes the output of each layer inside a block to every single layer that follows. These outputs are combined, allowing the network to leverage information from all previous layers.

However, connecting every layer to every other layer would be impractical. So, Densenet organizes these connections into groups called dense blocks. Within each block the number of connections grows continually according to the formula $\dfrac{N(N+1)}{2}$, where N is the number of layers in the block.

Densenet-121, used in this experiment, has four dense blocks with three transition layers in between. These transition layers help control the model's complexity by reducing the number of features before sending them to the next block. Finally, the last dense block connects to a classification layer for making predictions.\\
\textbf{Mobile-Net:} Slim Model with lightweight champion of deep learning \cite{howard2017mobilenets}. Unlike its bulkier models, Mobile-Net works on mobile devices where storing and training huge models is a no-go. Its secret weapon? Depthwise separable convolutions.

Imagine performing calculations across all colors in an image and doing this for each color separately, followed by a quick final touch-up (depthwise separable convolution). That's what Mobile-Net does, significantly reducing calculations without sacrificing much accuracy. Although, there exists a trade-off between a simple model (less complex)and accuracy.
\subsection{Datasets}
In our work, we utilize publicly available image datasets to evaluate the robustness of the pre-trained models. MNIST DATASET \cite{lecun1998mnist} contains images of handwritten digits with 60,000 training and 10,000 testing images containing ten classes (with numbers from 0 to 9). CIFAR10 and CIFAR100 datasets contain 60,000 32x32 color images(50,000 train and 10,000 tests), but CIFAR10 has ten different classes while CIFAR100 has 100. SVHN datasets obtained from house numbers in Google Street View images \cite{yuval2011reading} have ten classes with 73257 digits for training and 26032 digits for testing. 
\subsection{Attacks and Evaluation}
\textbf{Fast Gradient Sign Method(FGSM)}
powerful white-box adversarial attack used to generate adversarial examples. It involves calculating the gradient of the loss function with respect to the input, indicating the direction of maximum loss increase. Finally, add a scaled version of the gradient to the input \cite{goodfellow2014explaining}. FGSM aims to maximize the loss with respect to the true label that is:
\[
\max_{x^\prime \in \beta_\epsilon (x)} L(\theta; x^\prime
, y) = \max_{x^\prime \in \beta_\epsilon (x)} \mathcal{L}_{\text{ce}}(h_\theta(x^\prime
), y)
\]

\textbf{Projected Gradient Descent attack(PGD)} is a powerful white-box adversarial attack that leverages gradient descent to craft adversarial examples. These artificially perturbed inputs aim to fool machine learning models, causing them to misclassify with high confidence. It is a variant of FGSM that instead iteratively adds small noise to the image using the gradient\cite{madry2017towards}.\\
\textbf{Random Uniform Gaussian} is a basic white box attack that attempts to fool a machine learning model by adding carefully crafted random Gaussian noise to input images.

\section{Results}
\begin{figure}[h!]
  \centering
  \includegraphics[width=1.0\textwidth]{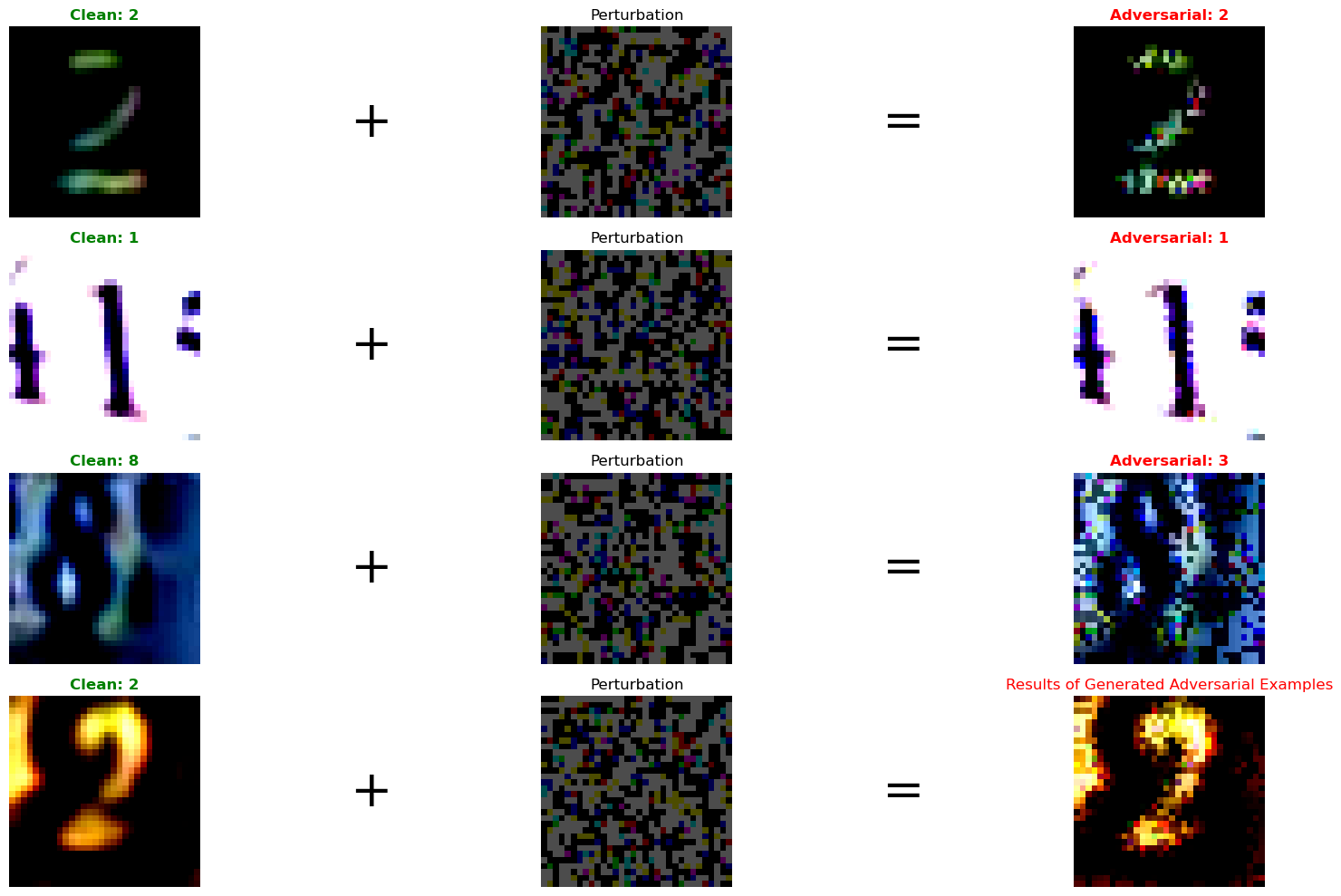}
  \caption{SVHN data with FGSM attack}
  \label{svhn}
\end{figure}

\begin{figure}[h!]
  \centering
  \includegraphics[width=1.0\textwidth]{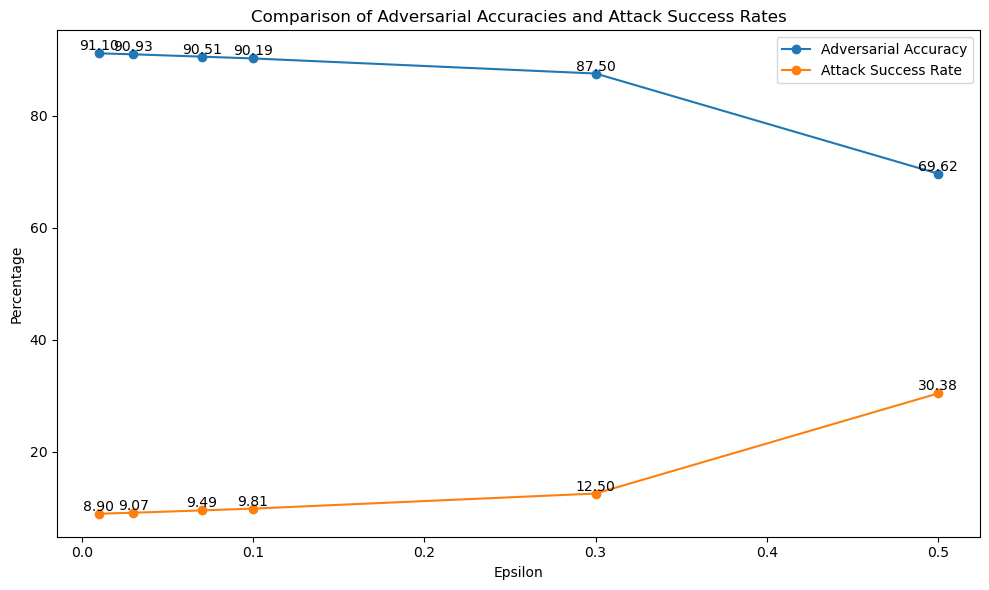}
  \caption{Mobile-net with PGD attack for MNIST data Results}
  \label{mnist}
\end{figure}

From our analysis, we present our accuracy on clean data, minimum and maximum adversarial accuracy after adversarial perturbation results in the tables \ref{clean_data}\ref{RUG}\ref{fgsm}\ref{pgd}. We also present a sample image after adversarial perturbation in figure \ref{svhn}. We also show a graphical representation with attack success rate and adversarial accuracy in the visualization\ref{mnist}.

After adding FGSM adversarial perturbation to the SVHN perturbation, we can visually see  from \ref{svhn}the change and in search of a circumstance pre-trained models might misclassify them.\\
Visualizing our mobile-net results for MNIST data after the PGD attack, we can clearly see that as the epsilon value increases, adversary accuracy reduces as the attack success rate increases. This implies that larger values of the adversarial perturbation control value epsilon also affect the adversarial accuracy, and therefore selection of epsilon is also key.\\
\begin{table}[h!]
 \centering
    \begin{tabular}{lcccr}
      \textit{Pretrained Model} & \textit{CIFAR10} & \textit{CIFAR100} & \textit{MNIST} &\textit{SVHN} \\\hline
      Dense-Net  &  \\
      
      Accuracy&  88.37 &66.05 &\color{blue} \textbf{99.66}& 93.88\\
            Epoch with best Results & 60 & 56& 17&54\\
            
       \hline
      Mobile-net&   &  & \\
 Accuracy &  86.20 &61.09 & 99.49& 93.42\\
            Epoch with best Results& 55 & 59 & 41&53\\
      \hline
      Resnet &   &  &  \\
 Accuracy&  \color{blue}92.442 &\color{blue}73.18 & 99.50& \color{blue}95.26\\
            Epoch with best Results & 39 & 51 & 58&59\\
      \hline
      
    \end{tabular}
    \caption{Evaluation on Clean Model}
    \label{clean_data}
\end{table}
   
\vspace{30pt}

    \centering
    \def\arraystretch{1.3}\footnotesize
    \begin{table}[h!]
    \centering
    \begin{tabular}{lcccr}
      \textit{Pretrained Model} & \textit{CIFAR10} & \textit{CIFAR100} & \textit{MNIST} &\textit{SVHN} \\\hline
      Dense-Net   &  \\
      
      Minimum Adv Accuracy &  17.24 & 4.79 & 24.68&26.82\\
            Maximum Adv Accuracy  & 50.76  & 21.24& 76.34&68.17\\
       \hline
      Mobile-net&   &  & \\
      Minimum Adv Accuracy&  32.73 & 10.97& 90.13&57.65\\
            Maximum Adv Accuracy  &  43.09 & 17.79 &\color{blue} \textbf{92.25}&69.49\\
      \hline
      Resnet &   &  &  \\
      Minimum Adv Accuracy& 40.32 & 12.30 &7.31 &65.09\\
            Maximum Adv Accuracy & \color{blue}51.15  & \color{blue}24.10 & 73.63& \color{blue}71.68\\
      \hline
      
    \end{tabular}  
      \caption{Random Uniform Gaussian Attack Case}
    \label{RUG}
\end{table}
    
\vspace{30pt}
    \centering
    \def\arraystretch{1.3}\footnotesize
    \begin{table}[h!]
        \centering
    \begin{tabular}{lcccr}
      \textit{Pretrained Model} & \textit{CIFAR10} & \textit{CIFAR100} & \textit{MNIST} &\textit{SVHN} \\\hline
      Dense-Net  &  \\
      
      Minimum Adv Accuracy&  17.74 &4.79  &33.24 & 26.82\\
            Maximum Adv Accuracy & \color{blue}50.79  & 21.24 & 78.24&68.17\\
       \hline
      Mobile-net &   &  & \\
      Minimum Adv Accuracy & 12.89  & 3.49& 10.22&20.15\\
            Maximum Adv Accuracy  &  36.81 & 17.82 & \color{blue} \textbf{96.06}&67.94\\
      \hline
      Resnet&   &  &  \\
      Minimum Adv Accuracy &  23.02& 7.67 & 36.94&36.20\\
            Maximum Adv Accuracy  &48.22   & \color{blue}24.02 & 77.44&\color{blue} 68.29\\
      \hline
      
    \end{tabular}    
          \caption{Fast Gradient Sign Method(FGSM)Attack Case}
    \label{fgsm}
\end{table}
    \vspace{30pt}
\begin{table}[h!]
        \centering

    \def\arraystretch{1.3}\footnotesize
    \begin{tabular}{lcccr}
      \textit{Pretrained Model} & \textit{CIFAR10} & \textit{CIFAR100} & \textit{MNIST} &\textit{SVHN} \\\hline
      Dense-Net&  \\
      
      Minimum Adv Accuracy&17.60   &  5.59& 21.23&28.28\\
            Maximum Adv Accuracy& 48.71  & 21.70 & 69.23&67.82\\
       \hline
      Mobile-net&   &  & \\
      Minimum Adv Accuracy & 14.35  & 4.10 & 69.62&22.80\\
            Maximum Adv Accuracy &  41.65 & 17.02& \color{blue} \textbf{91.10}&\color{blue}67.90\\
      \hline
      Resnet &   &  &  \\
      Minimum Adv Accuracy & 23.20 & 6.53 & 31.89&34.18\\
            Maximum Adv Accuracy & \color{blue} 49.39 & \color{blue}22.36  & 58.98& 66.41\\
      \hline
      
    \end{tabular} 
              \caption{Projected Gradient Descent attack(PGD) Attack Case}
    \label{pgd}
\end{table} 

\justifying

Our results in the tables show the performance of pre-trained models after fine-tuning, specifically Resnet, Dense-Net, and Mobile-net, across the datasets CIFAR10, CIFAR100, MNIST, and SVHN considering PGD, Random Uniform Gaussian and FGSM attacks. We evaluate based on the minimum and maximum adversarial accuracy under each attack, the clean model accuracy, and the epoch with the best results.\\
From table \ref{clean_data} of results and the evaluation on clean data, Dense-Net, the accuracy ranges from 66.05\% to 99.66\% across different datasets, with the epoch yielding the best results varying from 17 to 60. For Mobile-net, the accuracy ranged from 68.17\% to 90.13\%, and the epoch with the best results varied from 41 to 59. Resnet achieves an accuracy between 73.05\% and 99.50\% in the different datasets. Table \ref{RUG},\ref{fgsm} and \ref{pgd} of evaluation of different models, attacks, and datasets, Dense-Net achieved a minimum adversarial accuracy ranging from 4.79\% to 33.24\% across different datasets and attacks, with a maximum adversarial accuracy ranging from 21.24\% to 78.24\%.Mobile-net also demonstrated a minimum adversarial accuracy ranging from 3.49\% and 90.13\% and a maximum adversarial accuracy ranging from 17.02\% to 96.06\%. Finally, Resnet achieves a minimum adversarial accuracy ranging from 6.53\% to  65.09\% and a maximum adversarial accuracy ranging from 22.36\% to 77.44\%. Generally, from our results, Resnet50  demonstrates high accuracy on clean data for 60 epochs, particularly in the MNIST dataset when compared to Dense-Net. However, it is important to note that dataset variations significantly affect the model's performance. Additionally, Resnet50 shows high adversary accuracy with CIFAR10 and CIFAR100 datasets, while Mobile-net performs well on the MNIST dataset in almost all attacks. The adversarial accuracy of Resnet50 and Mobile-net decreases with smaller dataset samples. These findings suggest that Resnet50 and Mobile-net exhibit robustness in different data distributions and sample sizes and against various attacks. However, it's worth knowing that selecting a robust model for image classification requires understanding the distribution, size of the dataset, and computational resources required.\\

We achieved above 80\% of our initial plan, and therefore, to a greater extent, our research project was fruitful. Our main key takeaway was that the Robustness of fine-tuned pre-trained models varies across different data distributions and sample sizes. Therefore, in case of a need to make a decision on a robust pre-trained model for the image classification model, we will need to understand our data.\\
We had a computational resources problem with fine-tuning, therefore we were not able to evaluate and fine-tune all models we expected at the beginning, but generally, no underlying issues from our study need to be addressed urgently.


\justifying
\section{Conclusion and Reflections}
\subsection{Conclusion} In this project, we systematically evaluate the robustness of fine-tuned models considering different datasets and different attacks. Our result plays a crucial guide for researchers aiming to select pre-trained models tailored for image classification tasks while prioritizing robustness against potential attacks in their fine-tuned models. Our results show that ResNet50 demonstrates exceptional accuracy using 60 epochs on clean data, and it consistently exhibits high adversary accuracy on MNISt when challenged on CIFAR10 and CIFAR100. MobileNet excels well in defending against attacks on the MNIST dataset within different attacks. In essence, the robustness of fine-tuned models varies depending on factors such as data distribution, sample size, and the nature of the attacks encountered, indicating the need for tailored selection based on specific task requirements. 

\subsection{Reflections}
While it is always assumed that fine-tuning pre-trained models for image classification is helpful in saving computational resources, the computational resources are challenging when fine-tuning some of the heavy models in case of large datasets and insufficient computational power, and this is one of the challenges we faced in our work. If we were to redo this project, we would ensure we had enough computational resources to explore more fine-tuning possibilities. With a diverse background in our team, we did not find difficulties implementing this project. We discovered that comprehending and executing attacks was more manageable than we had anticipated initially. Hypothetically, our work can guide researchers and research organizations on the robustness of pre-trained models in image classification. We therefore hope to clean our models further, fine-tuning them with other techniques, explore other pre-trained models, and publish them.




 \bibliographystyle{abbrv}
\bibliography{reference.bib}

\end{document}